\documentclass[letterpaper, 10 pt, conference]{ieeeconf}  

\IEEEoverridecommandlockouts                              

\usepackage{microtype}
\usepackage{graphicx}
\usepackage{booktabs} 

\usepackage{hyperref}

\usepackage{amssymb}
\usepackage{amsmath}
\usepackage{amsthm}
\usepackage{mathtools}

\DeclareMathOperator*{\argmax}{arg\,max}

\usepackage[capitalize,noabbrev]{cleveref}

\usepackage{algorithm}
\usepackage{algorithmic}

\usepackage{caption}
\usepackage{subcaption}

\newtheorem{assumption}{Assumption}
\usepackage{xcolor}

\theoremstyle{plain}
\newtheorem{theorem}{Theorem}[section]
\newtheorem{proposition}[theorem]{Proposition}

\theoremstyle{definition}

\theoremstyle{remark}

\definecolor{darkred}{RGB}{139, 0, 0}
\definecolor{grey}{RGB}{137, 137, 137}

\title{Efficient and Generalizable Environmental Understanding for Visual Navigation}

\author{Ruoyu Wang$^{1}$, Xinshu Li$^{2}$, Chen Wang$^{3}$ and Lina Yao$^{4}$
\thanks{$^{1}$ Ruoyu Wang is with University of New South Wales, Sydney, Australia
        {\tt\small ruoyu.wang5@unsw.edu.au}}%
\thanks{$^{2}$ Xinshu Li is with University of New South Wales, Sydney, Australia
        }%
\thanks{$^{3}$ Chen Wang is with CSIRO's Data61, Sydney, Australia
        }%
\thanks{$^{4}$ Lina Yao is with CSIRO's Data61 and University of New South Wales, Sydney, Australia
        }%
}

\begin{document}

\maketitle
\thispagestyle{empty}
\pagestyle{empty}

\begin{abstract}

Visual Navigation is a core task in Embodied AI, enabling agents to navigate complex environments toward given objectives. Across diverse settings within Navigation tasks, many necessitate the modelling of sequential data accumulated from preceding time steps. While existing methods perform well, they typically process all historical observations simultaneously, overlooking the internal association structure within the data, which may limit the potential for further improvements in task performance. We address this by examining the unique characteristics of Navigation tasks through the lens of causality, introducing a causal framework to highlight the limitations of conventional sequential methods. Leveraging this insight, we propose Causality-Aware Navigation (CAN), which incorporates a Causal Understanding Module to enhance the agent's environmental understanding capability. Empirical evaluations show that our approach consistently outperforms baselines across various tasks and simulation environments. Extensive ablations studies attribute these gains to the Causal Understanding Module, which generalizes effectively in both Reinforcement and Supervised Learning settings without computational overhead.

\end{abstract}

\section{Introduction}

Navigation aims to enable agents to reach specified locations based on given instructions or objectives. Its methods can be broadly classified into {\it Supervised Learning methods} and {\it Reinforcement Learning methods}, differentiated by their training processes \cite{francis2022core}. Supervised methods learn from demonstrations, seeking to replicate the demonstrator's behavior, while Reinforcement Learning (RL) trains agents to maximize rewards through interactions with the environment.


In both settings, despite the advancements in recent years, many approaches share common limitations. Existing methods typically rely on traditional sequential data modelling techniques such as RNN \cite{hochreiter1997long} and Transformer \cite{vaswani2017attention}, using historical observations and actions to learn the associations between them. While these strategies may appear effective, it is important to recognise that navigation tasks exhibit unique characteristics in their association structures from a causal perspective (Section~\ref{bkgd_motivation}-\ref{bkgd_causal_framework}). However, existing methods have not sufficiently addressed these distinctions, thereby limiting their effectiveness.

To address such limitations, we propose a novel approach in this paper. First, we analyze the disparities between Navigation tasks and other types of sequential tasks through the lens of causality (Section~\ref{bkgd_motivation}-\ref{bkgd_causal_framework}). By framing these distinctions within a causal framework, we highlight the shortcomings of conventional sequential data modeling methods in handling navigation tasks. Building on this, we propose an innovative solution incorporating a Causal Understanding Module, designed to enhance the model's environmental understanding capabilities in these contexts (Section~\ref{method_transformer}). Our framework avoids task-specific architectures or inductive biases, ensuring generalizability and reproducibility, and can be trained end-to-end, offering a flexible and adaptable solution aligned with principles from \cite{khandelwal2022simple}. In summary, our contributions are threefold:

\begin{figure}
    \centering
    \includegraphics[width=0.75\linewidth]{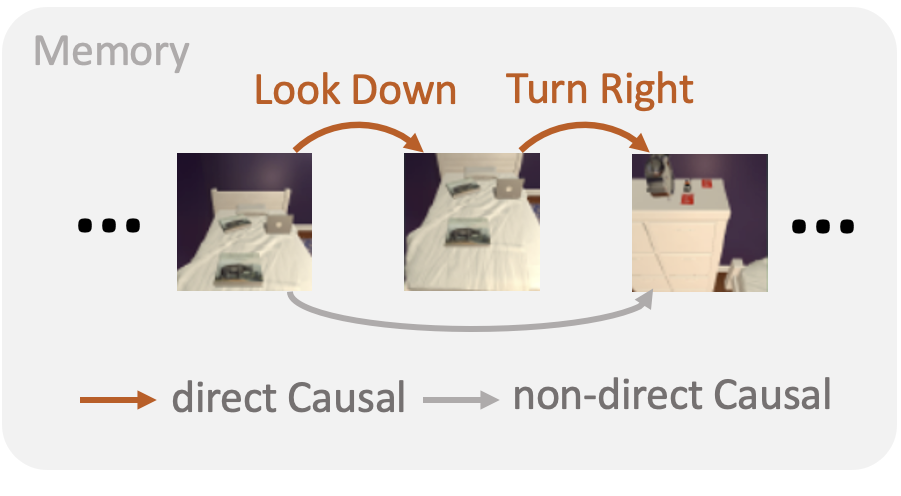}
    \caption{A key characteristic of Navigation is that \textcolor{darkred}{direct causal relationships} are typically single-step. To move from one state to another, an agent must traverse all intermediate states without skipping. Our method enhances environmental understanding by prioritizing these \textcolor{darkred}{direct causal associations} over \textcolor{grey}{other associations.}}
    \label{fig:direct_causal}
  \vspace{-15pt}
\end{figure}

\begin{itemize}

    \item We propose Causality-Aware Navigation (CAN), a novel End-to-End framework for Navigation tasks without task-specific inductive bias, and experimentally show that it outperforms the baseline methods by a significant margin across various tasks and simulators.

    \item We introduce a causal framework tailored for Navigation, offering a comprehensive rationale for the limitations in existing methods that may impede agent performance. Based on this causal framework, we design a Causal Understanding Module to enhance the method's effectiveness. This enhancement is achieved without increasing computational overhead, showcasing the framework's ability to optimize performance through a deeper understanding of causal relationships.
    
    \item We conduct extensive ablation studies to evaluate the impact of the Causal Understanding Module, demonstrating its efficiency, effectiveness, and necessity. Additionally, while our primary focus is on navigation tasks in reinforcement learning, experiments reveal that the proposed method also delivers significant benefits in supervised learning settings.

\end{itemize}

\section{Background}
\label{bkgd}

\subsection{Problem Formulation}
\label{bkgd_formulation}
We model the agent's interaction with the environment as a Partially Observable Markov Decision Process (POMDP) \cite{aastrom1965optimal}. A POMDP provides a formal framework for situations where the agent cannot directly observe the true state of the environment but must act based on incomplete observations. A POMDP is formally defined by the tuple $(S, A, T, R, \Omega, O, \gamma)$, where $S$ is the set of states of the environment, $s \in S$ represents the true hidden state, $A$ is the set of actions the agent can take, $T: S \times A \times S \to [0, 1]$ is the transition function, with $T(s' \mid s, a)$ representing the probability of transitioning from state $s$ to state $s'$ after taking action $a$, $R: S \times A \to \mathbb{R}$ is the reward function, where $R(s, a)$ denotes the immediate reward received after taking action $a$ in state $s$, $\Omega$ is the set of possible observations, $O: S \times A \times \Omega \to [0, 1]$ is the observation function, with $O(o \mid s', a)$ being the probability of observing $o$ after transitioning to state $s'$ due to action $a$, $\gamma \in [0, 1)$ is the discount factor, which controls the trade-off between immediate and future rewards. The goal of the agent is to find a policy $\pi$ that maximizes the expected cumulative reward:
\[
\max_{\pi} \mathbb{E} \left[ \sum_{t=0}^{\infty} \gamma^t R(s_t, a_t) \right],
\]
where actions $a_t$ are chosen according to the policy $\pi$, and $s_t$ is the hidden state at time $t$. In our problem of Navigation, the agent is required to finish a task with a given \textit{Objective}, By accomplishing this task, the agent obtains the reward $R$.

\subsection{Motivation}
\label{bkgd_motivation}

Across various navigation tasks, modeling sequential observation data from past time steps is crucial. Existing methods typically treat this as a sequential modeling problem, training models to associate past observations $\{ o_{i} \}_{i=0}^{t}$ and actions $\{ a_{i} \}_{i=0}^{t-1}$ with the current action $a_{t}$. However, we argue that this approach may constrain the agent's performance.

In particular, we observe differences in temporal dependency structures between navigation tasks and other sequential tasks. For instance, language models must capture long-term dependencies (e.g., syntax and discourse coherence), as evidenced by the effectiveness of attention mechanisms in existing literature \cite{vaswani2017attention}. In contrast, navigation tasks exhibit shorter-term dependencies: the next state is primarily determined by the current state and action. While historical states may influence future states in partially observable settings, these dependencies are often mediated through the current state (Figure~\ref{fig:direct_causal}). Our experiments in Section~\ref{exp_ablation} also quantitatively support this distinction, demonstrating that prioritizing shorter-term dependencies can significantly enhance navigation performance.

This distinction in dependency structures helps explain why traditional sequential models, such as RNNs and Transformers, may underperform in navigation tasks without architectural adaptation. These models are inherently designed to capture long-term dependencies through mechanisms like attention \cite{vaswani2017attention} or recurrent state propagation \cite{hochreiter1997long}, which are crucial for NLP tasks (e.g., resolving coreference across paragraphs). However, in navigation tasks, the next state is primarily determined by the immediate state-action pair. As a result, models that emphasize long-term dependencies may overparameterize the problem, leading to unnecessary computational complexity or overfitting to spurious correlations. To address this issue, we propose a novel causal framework that can be integrated into RNNs and Transformers {\textit without} additional computational overhead, enabling them to focus more effectively on short-term dependencies.

\begin{figure}
    \centering
    \includegraphics[width=0.7\linewidth]{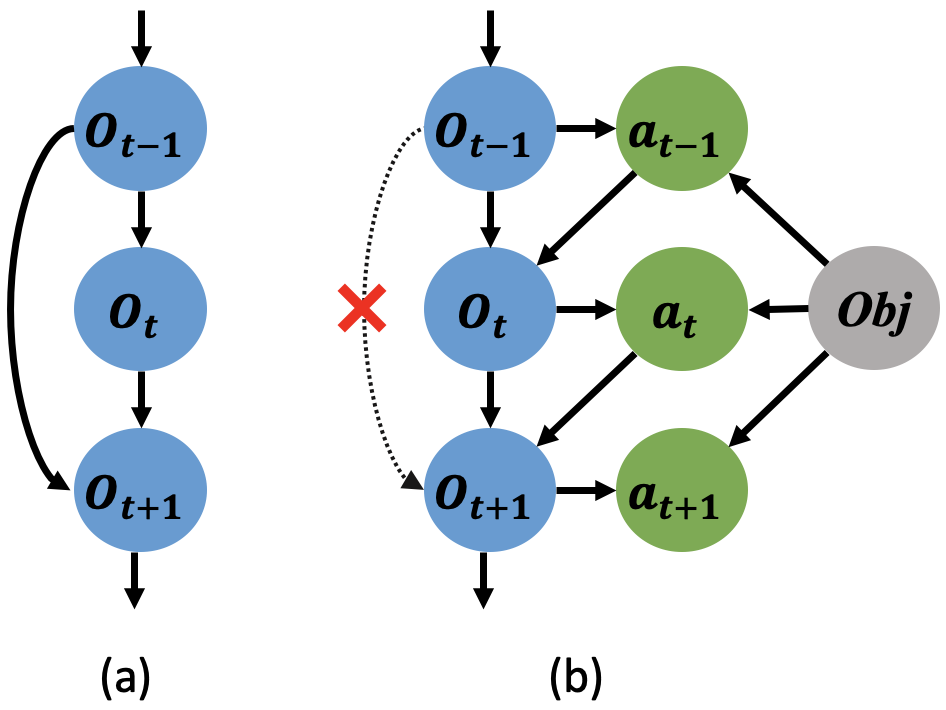}
    \caption{(a) Capturing long-term dependencies, such as $O_{t-1} \rightarrow O_{t+1}$, is a critical capability of efficient language models, as long-term direct causal relationships are present. (b) In Navigation tasks, long-term dependencies are weaker due to the absence of long-term direct causation.}
    \label{fig:causal_graph}
\end{figure}

\begin{figure*}
  \centering
  \includegraphics[width=0.8\linewidth]{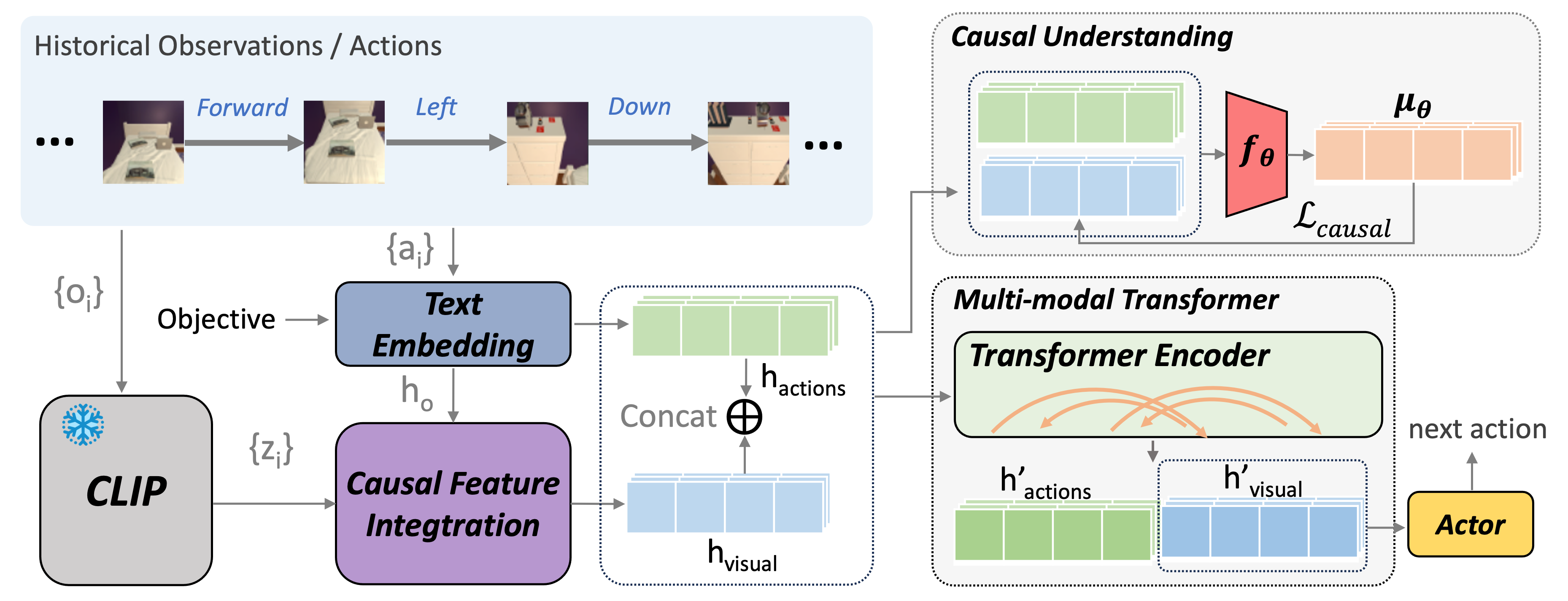}
  \caption{The workflow of our proposed method. First, we process the observations by the frozen CLIP vision model and process the objective and previous actions by simple Embedding modules. Then we integrate the feature of objective and observations by a Causal Feature Integration Module. We concatenate the features of the observations and the actions and pass these features to the Causal Understanding Module, which introduces an auxiliary objective that encourages the model to enhance its capacity for environmental understanding. Finally, the features are processed with a Multi-modal Transformer, and the Actor layer takes the post-processed visual features as input to predict the action at the current time step.}
  \label{fig:framework}
  
\end{figure*}

\subsection{Causal Framework Formulation}
\label{bkgd_causal_framework}

We formalize the idea in Section~\ref{bkgd_motivation} within a causal framework, starting with the following assumptions.

\begin{assumption}
    At any time step $t$, the observation $O_t$ and the action $a_t$ are the sole direct causal parents that determine the subsequent observation $O_{t+1}$.
    \label{assumption:state_action}
\end{assumption}

\begin{assumption}
    At any time step $t$, the observation $O_t$ and the Objective are the only direct causes of the action $a_t$.
    \label{assumption:state_goal_action}
\end{assumption}

Notably, while these assumptions may resemble the Markov property \cite{puterman1990markov}, they differ fundamentally. While Markov property concerns transition probabilities, our assumptions pertain to causality. As a result, our framework is inherently non-Markov, as discussed in Section~\ref{bkgd_formulation} and Proposition~\ref{proposition:no_markov}.

\begin{proposition}
    At any time step $t$, the conditional probability of transitioning to the next observation $O_{t+1}$ depends on the entire history of previous observations, i.e., $ P(O_{t+1} \mid O_{t} = o_{t}, \dots, O_{1} = o_{1}) \neq P(O_{t+1} \mid O_{t} = o_{t})$.
    \label{proposition:no_markov}
\end{proposition}

We depict these causal assumptions in a causal graph in Figure~\ref{fig:causal_graph}b, where the edge $O_{t-1} \rightarrow O_{t} \leftarrow a_{t-1}$ corresponds to Assumption~\ref{assumption:state_action}, and the edge $O_{t} \rightarrow a_{t} \leftarrow Obj$ corresponds to Assumption~\ref{assumption:state_goal_action}. Further, we have Proposition~\ref{proposition:no_long_term} based on the assumptions, suggesting that there exist NO direct causal relationships between $O_{t-1}$ and $O_{t+1}$, so $O_{t-1} \rightarrow O_{t+1}$ in Figure~\ref{fig:causal_graph}b is marked by a light dotted line, indicating the causation between them is indirect and relatively weaker.

\begin{proposition}
    At any given time step $t$, and for any integer $\delta \geq 2$, there exist no direct causal relationships between $O_{t}$ and $O_{t-\delta}$, the causal relationships between observations $O_{t}$ and $O_{t-\delta}$ are indirect and must be mediated by $O_{t'}$ for all $t'$ where $t-\delta \leq t' \leq t$.
    \label{proposition:no_long_term}
\end{proposition}

As discussed in Section~\ref{bkgd_motivation}, Proposition~\ref{proposition:no_long_term} distinguishes Navigation tasks from other sequential data tasks. Thus, we aim to emphasize the direct causal relationships $O_{t-1} \rightarrow a_{t-1}$, $O_{t-1} \rightarrow O_{t}$, and $a_{t-1} \rightarrow O_{t}$ in our framework. While $O_{t-1} \rightarrow a_{t-1}$ and $O_{t-1} \rightarrow O_{t}$ are inherently highlighted by the data collection process, $a_{t-1} \rightarrow O_{t}$ requires further attention. To address this, we have Proposition~\ref{proposition:d-seperation}, which specifies the condition for this edge’s existence.

\begin{proposition}
    At any given time step $t$, let $G_{t}$ denote the local causal graph induced by $O_{t}=o$. The causal relationship $a_{t-1} \rightarrow O_{t}$ exist in $G_{t}$ under the intervention $do(a_{t-1}:\pi (a|o_{t-1}))$ if and only if $O_{t}\not\!\perp\!\!\!\perp a_{t-1} | O_{t-1}$.
    \label{proposition:d-seperation}
\end{proposition}

With Proposition~\ref{proposition:d-seperation}, the causal association between $a_{t-1}$ and $O_{t}$ can be measured by Conditional Mutual Information \cite{cover1999elements,seitzer2021causal}, as derived below.
\begin{equation}
\label{eq:cmi}
    \begin{array}{l}
    I(O_{t};a_{t-1}|O_{t-1}) \\[1ex]
    = \int_{O_{t-1}}D_{KL}(P_{O_{t},a_{t-1}|O_{t-1}} || P_{O_{t} | O_{t-1}} \otimes P_{a_{t-1} | O_{t-1}} ) dO_{t-1} \\[1ex]
    = \mathbb{E}_{a_{t-1}|O_{t-1}} D_{KL}(P_{O_{t}|O_{t-1}, a_{t-1}} || P_{O_{t} | O_{t-1}})\nonumber
    \end{array}
\end{equation}

Building on these concepts, we propose a method to estimate this quantity and distinguish it from other associations. The implementation details are elaborated in Section~\ref{method_transformer}.

\begin{table*}[t]
    \centering
    \caption{Comparison of results across RoboTHOR ObjNav (Left), Habitat PointNav (Middle), and Habitat ObjNav (Right). EmbCLIP results are obtained as described in Section~\ref{exp_setting} for a fair comparison, while results for other methods are sourced from the top-performing entries on each task's leaderboard. Overall, our method significantly outperforms existing approaches. SR and GD evaluations are unavailable for some prior works due to the lack of publicly available record.}
    \renewcommand{\arraystretch}{1}
    \begin{minipage}{0.27\linewidth}
        \centering
        \small
        \label{tab:rst_objnav_robothor}
        \medskip
        \begin{tabular}{@{}l|cc@{}} 
            \toprule
            & SPL $\uparrow$ & SR $\uparrow$ \\
            \midrule
            CAN (Ours) & \textbf{0.31{\scriptsize$\pm$.03}} & \textbf{0.73{\scriptsize$\pm$.05}} \\
            Emb-CLIP & 0.18{\scriptsize$\pm$.02} & 0.42{\scriptsize$\pm$.03} \\
            \midrule
            ProcTHOR  & 0.27 & 0.66 \\
            Action Boost & 0.17 & 0.37 \\
            RGB+D & 0.17 & 0.35 \\
            ICT-ISIA & 0.18 & 0.38 \\
            \bottomrule
        \end{tabular}
    \end{minipage}
    \hfill
    \begin{minipage}{0.35\linewidth}
        \centering
        \small
        \label{tab:rst_pointnav_habitat}
        \medskip
        \begin{tabular}{@{}l|ccc@{}}
            \toprule
            & SPL $\uparrow$ & SR $\uparrow$ & GD $\downarrow$ \\
            \midrule
            CAN (Ours) & \textbf{0.93{\scriptsize$\pm$.02}} & \textbf{0.98{\scriptsize$\pm$.03}} & \textbf{0.35{\scriptsize$\pm$.01}} \\
            Emb-CLIP & 0.84{\scriptsize$\pm$.02} & 0.95{\scriptsize$\pm$.02} & 0.46{\scriptsize$\pm$.01} \\
            \midrule
            DD-PPO & 0.89 & - & - \\
            MDP & 0.85 & - & - \\
            Arnold & 0.70 & - & - \\
            SRK AI Lab & 0.66 & - & - \\
            \bottomrule
        \end{tabular}
    \end{minipage}
    \hfill
    \begin{minipage}{0.35\linewidth}
        \centering
        \small
        \label{tab:rst_objnav_habitat}
        \medskip
        \begin{tabular}{@{}l|ccc@{}}
            \toprule
            & SPL $\uparrow$ & SR $\uparrow$ & GD $\downarrow$ \\
            \midrule
            CAN (Ours) & \textbf{0.16{\scriptsize$\pm$.01}} & \textbf{0.39{\scriptsize$\pm$.02}} & \textbf{6.64{\scriptsize$\pm$.18}} \\
            Emb-CLIP & 0.07{\scriptsize$\pm$.01} & 0.16{\scriptsize$\pm$.02} & 8.03{\scriptsize$\pm$.21} \\
            \midrule
            RIM & 0.15 & 0.37 & 6.80 \\
            PIRLNav & 0.14 & 0.35 & 6.95 \\
            Stubborn & 0.10 & 0.22 & 9.17 \\
            TreasureHunt & 0.09 & 0.21 & 9.20 \\
            \bottomrule
        \end{tabular}
    \end{minipage}
    \label{tab:three_tables}
\end{table*}

\section{Causality-Aware Navigation}
\label{method_transformer}

The architecture of our method is depicted in Figure~\ref{fig:framework}. The functions and operations of each individual modules are delineated in the subsequent sections.

\textbf{Visual Encoder} We use the CLIP ResNet-50 model \cite{radford2021learning} for Visual Perception to process the observation $o_{i}$ into features $z_{i}$, as CLIP has been shown to performed well for Embodied AI tasks \cite{khandelwal2022simple}. The Visual Encoder remains fixed during training.

\textbf{Text Embedding} The Objective and Actions are given as text, e.g., \textit{Find the laptop}, \textit{Move Forward}. Since tasks have a finite set of targets and actions, we employ a simple Embedding layer to map the Objective and Actions $a_i$ to features $h_o$ and $h_{a_i}$.

\textbf{Causal Feature Integration} After extracting visual features $z_{i}$ and the goal representation $h_{o}$, we process them through the Causal Feature Integration Module. This module serves two purposes: 1) aligning with Assumption~\ref{assumption:state_goal_action}, it integrates objective-related information into the visual representation, reinforcing the causal relationship between them; and 2) refining CLIP features through a tunable layer to enhance performance. Consequently, it generates a distinct feature set $h_{o_{i}}$ for each observation.

\textbf{Multi-modal Transformer} The features from the previous steps are passed into the Multi-modal Transformer. For simplicity, we denote the sequences $\{h_{o_{i}}\}$ and $\{h_{a_{i}}\}$ as $h_{visual}$ and $h_{action}$, respectively. These are concatenated along the time-step dimension (Equation~\ref{eq:concat}), and Positional Encoding is applied to $h_{concat}$. The features are then processed through a transformer encoder \cite{vaswani2017attention} with a causal attention mechanism to prevent attending to future time steps. The transformer outputs an updated feature set $h'_{concat}$.
\begin{equation}
    h_{concat} = \left[ h_{visual}; h_{actions} \right]
    \label{eq:concat}
\end{equation}
Similarly, the transformer output contains updated features for the visual states and actions, as shown in Equation~\ref{eq:output_concat}. The updated visual features $h'_{visual}$ are then passed to an Actor layer to predict the action for the given state (Equation~\ref{eq:actor}).
\begin{equation}
    h'_{concat} = \left[ h'_{visual}; h'_{actions} \right]
    \label{eq:output_concat}
\end{equation}
\begin{equation}
    a_{t} = Actor(h'_{visual})
    \label{eq:actor}
\end{equation}

\textbf{Causal Understanding Module} Following the idea in Section~\ref{bkgd_motivation} - \ref{bkgd_causal_framework}, we encourage the one-step direct causation to stand out from the associations in other forms. Formally, we construct a causal loss building upon the estimation process of the conditional mutual information term in Section~\ref{bkgd_causal_framework}. First, we expand this term:
\begin{multline*}
I(O_{t};a_{t-1}|O_{t-1}) \\
= \mathbb{E}_{a_{t-1}|O_{t-1}} \mathbb{E}_{O_{t}|O_{t-1}, a_{t-1}} \left[ \log \frac{P(O_{t}|O_{t-1},a_{t-1})}{P(O_{t}|O_{t-1})} \right] \nonumber
\end{multline*}
With the normality assumption, we estimate the distribution $P(O_{t}|O_{t-1},a_{t-1})$ by a probabilistic neural network $f_{\theta}$, i.e., to parametrize it as a Gaussian Distribution $\mathcal{N}(\mu_{\theta}(o_{t-1}, a_{t-1}), \sigma^{2}_{\theta}(o_{t-1}, a_{t-1}))$, where $\mu_{\theta}$, $\sigma^{2}_{\theta}$ are the outputs of the network $f_{\theta}$. Over the data collection process, a set of data $\mathcal{D}=\{ (o^{(i)}, a^{(i)}, o^{(i+1)}) \}_{i=1}^{N}$ from the joint distribution are accumulated, so the network can be estimated by Maximum Likelihood Estimation on $\mathcal{D}$:
\begin{equation}
\label{eq:nll}
    \theta^{*} = \argmax_{\theta} \left( \frac{1}{\sqrt{2\pi\sigma^{2}_{\theta}}} \right) \exp \biggl\{ -\sum_{i=1}^{N} \frac{(o_{i}-\mu_{\theta})^2}{2\sigma_{\theta}^2} \biggl\}
\end{equation}
Further, the distribution $P(O_{t} | O_{t-1})$ can be decomposed to:
\begin{equation}
    P(O_{t} | O_{t-1}) = \int_{a_{t-1}} P(O_{t}|O_{t-1},a_{t-1}) P(a_{t-1}|O_{t-1}) da_{t-1}  \nonumber
\end{equation}
This term can be estimated by Monte-Carlo Sampling from $P(O_{t} | O_{t-1}, a_{t-1})$. Specifically, we randomly sample $K$ points from the datasaet $\mathcal{D}$, then it can be estimated by:
\begin{equation}
    P(O_{t} | O_{t-1}) \approx \frac{1}{K} \sum_{k=1}^{K} P(O_{t} | O_{t-1}, a^{(k)})
\end{equation}
This means $P(O_{t} | O_{t-1}, a_{t-1})$ can be seen as a Mixture of Gaussian Distribution, and our objective of the Mutual Information can be written as below:
\small
\begin{align}
    I(O_{t};a_{t-1}|O_{t-1}) & = \mathbb{E}_{a_{t-1}|O_{t-1}} D_{KL}(P_{O_{t}|O_{t-1}, a_{t-1}} || P_{O_{t} | O_{t-1}}) \nonumber \\  
    & \approx \frac{1}{K} \left[ D_{KL}(f_{\theta}(O_{t-1}, a_{t-1}) | g_{\theta}(O_{t-1}, a_{t-1}) \right]  \nonumber
\end{align}
\normalsize
where $g_{\theta} = \frac{1}{K} \sum_{k=1}^{K} f_{\theta}(O_{t}, a^{(k)})$. Our objective thus becoming to maximize the KL divergence between $f_{\theta}$, a Gaussian distribution, and $g_{\theta}$, a Mixture of Gaussians. Since directly computing this KL divergence is intractable, we adopt its lower and upper bounds from \cite{durrieu2012lower}, estimate it using the midpoint of these bounds, and use this estimator as the objective for our Causal Understanding Module. In practice, we implement a causal loss as defined in Equation~\ref{eq:loss_causal}: 
\begin{equation}
    \mathcal{L}_{causal}(\theta) = \mathbb{E}_{t} \left[ (\mu_{\theta} - h_{v_{t+1}} )^2 \right]
    \label{eq:loss_causal}
\end{equation}
This loss term closely aligns with accurately predicting the next state given the current state and action, reinforcing both the theoretical foundation and practical intuition of our approach. An ideal agent with sufficient environmental knowledge should predict \( O_{t+1} \) from \( O_t \) and \( a_t \) accurately. Prior works have explored similar ideas in estimating the causal effect of actions on states, primarily in the context of sample efficiency \cite{seitzer2021causal, sontakke2021causal}, or leveraging next-state prediction for rewards in RL \cite{pathak2017curiosity} and planning \cite{wang2018look}. In contrast, our method introduces a causal loss, providing a principled framework to explain how it enhances learning, marking a key contribution of our work.

\textbf{Training Process} For experiments in the Reinforcement Learning setting, we train our model with Proximal Policy Optimization (PPO) \cite{schulman2017proximal}. So the overall objective of our framework becomes Equation~\ref{eq:loss_overall}, where $\mathcal{L}_{PPO}$ denotes the original objective of PPO \cite{schulman2017proximal} which is to be maximized, $\mathcal{L}_{causal}$ is our proposed Causal Loss which is to be minimized, so we subtract the causal loss from the PPO objective.
\begin{equation}
    \mathcal{L}_{total}(\theta) = \mathcal{L}_{PPO} - \alpha \mathcal{L}_{causal}
    \label{eq:loss_overall}
\end{equation}
We also conduct experiments in the Supervised Learning setting, where the training process differs from above, which will be elaborated in Section~\ref{exp_other_setting}.



\section{Experiments}
\label{exp}


\subsection{Experiment Setting}
\label{exp_setting}

\textbf{Task Descriptions}
We evaluate our method on three tasks in the Reinforcement Learning setting:
(1) \textbf{Object Navigation in RoboTHOR} \cite{deitke2020robothor}, where an agent navigates to find a specified object (e.g., ``Find an apple''). The task includes 12 object categories, and the agent can \textit{MoveAhead}, \textit{RotateRight}, \textit{RotateLeft}, \textit{LookUp}, and \textit{LookDown}. The task is complete when the agent performs \textit{Stop} with the target object visible within 1 meter.
(2) \textbf{Object Navigation in Habitat}, similar to RoboTHOR but with 21 object categories. Unlike RoboTHOR, the agent does not need to face the target object. Habitat uses real-world indoor scenes from MatterPort3D \cite{chang2017matterport3d}. 
(3) \textbf{Point Navigation in Habitat}, where an agent moves from a random start to specified polar goal coordinates (e.g., ``Navigate to (X, Y)''). The agent can \textit{MoveAhead}, \textit{RotateRight}, and \textit{RotateLeft}, and must perform \textit{Done} upon reaching the goal. Training occurs in the Gibson Database \cite{xia2018gibson}.

\textbf{Evaluation Metrics}  
We use multiple metrics to evaluate an agent's performance. For Object Navigation in RoboTHOR, we follow prior work and report Success Rate (SR) and Success weighted by Path Length (SPL). SR measures the proportion of successful tasks, while SPL accounts for both success and path efficiency. For the two Habitat tasks, we additionally consider Goal Distance (GD), which quantifies the agent's final distance to the goal. Higher SR and SPL or lower GD indicate better performance.

\begin{figure}
  \centering
  \begin{subfigure}{0.23\textwidth}
        \includegraphics[width=\textwidth]{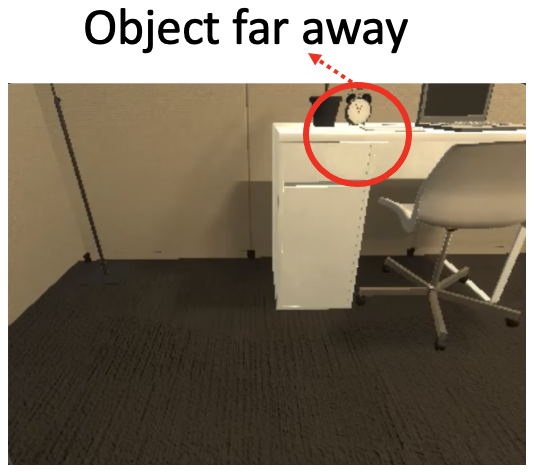}
        \caption{}
        \label{fig:case_eg_no_causal}
    \end{subfigure}
    \hfill
    \begin{subfigure}{0.23\textwidth}
        \includegraphics[width=\textwidth]{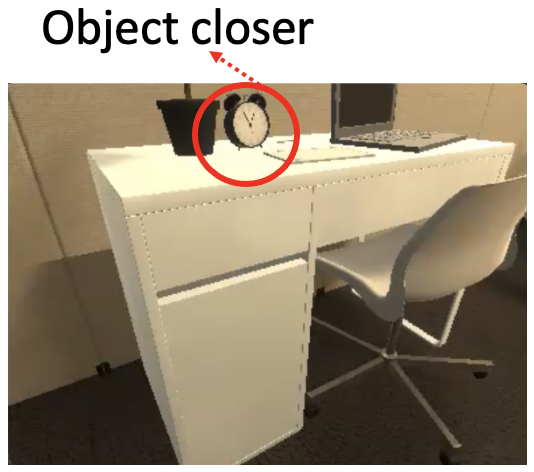}
        \caption{}
        \label{fig:case_eg_with_causal}
    \end{subfigure}
  \caption{Comparison on RoboTHOR ObjNav \textit{Find an AlarmClock} task. (a) EmbCLIP stops far away from the goal object, making it fail to complete the task. (b) Our method stops at a closer spot, thus benefiting the performance.}
  \label{fig:case_eg}
\end{figure}

\textbf{Baselines} 
We use EmbCLIP \cite{khandelwal2022simple} as our main baseline for several reasons: 1) Both methods share the same experimental setup, as neither incorporates task-specific designs and both are trained end-to-end, allowing a direct comparison. 2) Both use the same vision backbone and model architecture for key modules, ensuring fairness.  
3) EmbCLIP is open-sourced, enabling reproducibility and thorough analysis. In our experiments, both methods are trained for the same number of steps under identical settings. 

Beyond direct comparison with EmbCLIP, we also evaluate our method against task-specific approaches. Note these methods are NOT directly comparable as they incorporate strong task-specific inductive biases to enhance the model performance, such as dataset-specific hand-crafted logic, semantic mapping, additional large-scale datasets, or multi-stage training. While these enhancements improve task performance, they limit generalizability and reproducibility. Nonetheless, for completeness, we report their results in the corresponding tables. These methods include Action Boost, RGB+D ResNet18, ICT-ISIA, and ProcTHOR \cite{deitke2022} for RoboTHOR ObjNav; Stubborn \cite{luo2022stubborn}, TreasureHunt \cite{maksymets2021thda}, Habitat on Web (IL-HD) \cite{ramrakhya2022habitat}, Red Rabbit \cite{ye2021auxiliary}, PIRLNav \cite{ramrakhya2023pirlnav}, and RIM for Habitat ObjNav; and DD-PPO \cite{wijmans2019dd}, Monocular Predicted Depth (MDP), Arnold, and SRK AI for Habitat PointNav. The reported results are sourced from the official leaderboards. If certain metrics or citations are missing, it is due to the unavailability of the corresponding paper or code.


\begin{table*}
  \centering
  \caption{Result of Ablation Studies. The majority of the observed improvements in Section~\ref{exp_rst} can be attributed to the Causal Understanding Module. Removing this module results in a significant performance drop, while incorporating it into the baseline RNN module leads to a significant performance improvement across all tasks and metrics.}
  \setlength{\tabcolsep}{6pt} 
  \begin{tabular}{l | l l | l l l | l l l}
    \toprule
    & \multicolumn{2}{c}{RoboTHOR ObjNav} & \multicolumn{3}{c}{Habitat ObjNav} & \multicolumn{3}{c}{Habitat PointNav}\\
    & SPL $\uparrow$ & SR $\uparrow$ & SPL $\uparrow$ & SR $\uparrow$ & GD $\downarrow$ & SPL $\uparrow$ & SR $\uparrow$ & GD $\downarrow$ \\
    
    \midrule

    CAN & 0.31 & 0.73 & 0.16 & 0.39 & 6.64 & 0.93 & 0.98 & 0.35 \\
    
    \midrule
    
    Transformer & 0.19 \textcolor{teal}{$\downarrow$(-0.12)}  & 0.37 \textcolor{teal}{$\downarrow$(-0.36)} & 0.08 \textcolor{teal}{$\downarrow$(-0.08)}& 0.19 \textcolor{teal}{$\downarrow$(-0.19)}& 7.08 \textcolor{teal}{$\uparrow$(+0.34)}& 0.84 \textcolor{teal}{$\downarrow$(-0.09)}& 0.93 \textcolor{teal}{$\downarrow$(-0.05)}& 0.45 \textcolor{teal}{$\uparrow$(+0.10)} \\
    
    Causal-RNN & 0.25 \textcolor{teal}{$\downarrow$(-0.06)} & 0.54 \textcolor{teal}{$\downarrow$(-0.19)}& 0.13 \textcolor{teal}{$\downarrow$(-0.05)}& 0.29 \textcolor{teal}{$\downarrow$(-0.09)}& 6.98 \textcolor{teal}{$\uparrow$(+0.24)}& 0.90 \textcolor{teal}{$\downarrow$(-0.03)} & 0.96 \textcolor{teal}{$\downarrow$(0.02)} & 0.40 \textcolor{teal}{$\uparrow$(+0.05)} \\
    
    
    
    \bottomrule
  \end{tabular}
  \label{tab:ablation_rst}
\end{table*}

\textbf{Environment and Parameters} 
For all experiments, we trained our method for 100M steps. During training, we selected the checkpoint with the highest Success Rate (SR) for evaluation on other metrics. Regarding the model architecture, we employ an Embedding layer for Action and Objective Embedding, and a linear layer for Feature Merging, the Causal Understanding Module, and the Actor Module. The Multi-modal Transformer Encoder consists of a single-layer transformer encoder with 4 self-attention heads, each with a dimension of 392, resulting in a total dimension of 1568. The episode length is set to 128. The weight for the Causal Loss $\alpha$ is 1. We train our framework using Adam \cite{kingma2014adam} with a learning rate of $1 \times 10^{-4}$, which is scheduled to linearly decay to 0 by the end of training. Our implementation is based on AllenAct \cite{weihs2020allenact}, a learning framework designed for Embodied-AI research. Any settings not explicitly specified here remain consistent with \cite{khandelwal2022simple}. All experimental results presented in this paper adhere to these settings unless otherwise specified.

\begin{figure}
  \centering
  \begin{subfigure}{0.23\textwidth}
        \includegraphics[width=\textwidth]{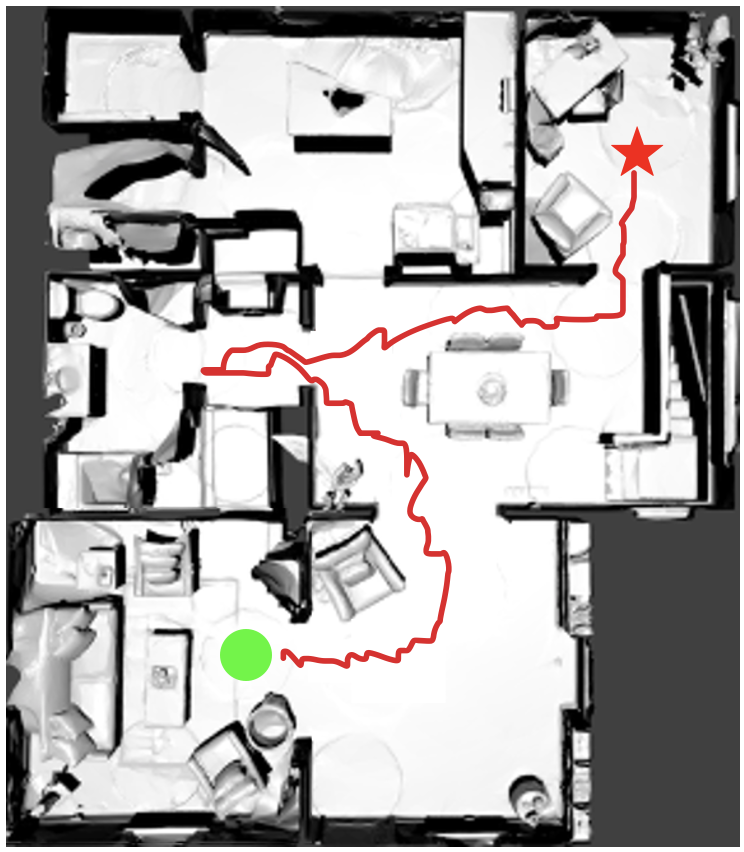}
        \caption{}
        \label{fig:case_eg_no_causal_habitat}
    \end{subfigure}
    \hfill
    \begin{subfigure}{0.23\textwidth}
        \includegraphics[width=\textwidth]{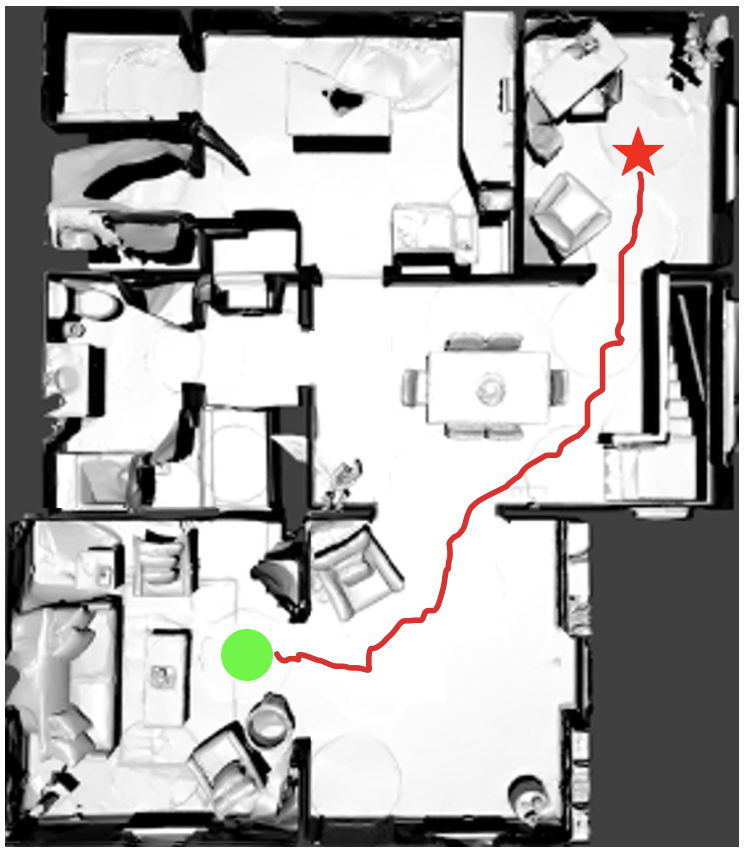}
        \caption{}
        \label{fig:case_eg_with_causal_habitat}
    \end{subfigure}
  \caption{Comparison of our method and EmbCLIP on Habitat PointNav. The green dot marks the start, and the red star marks the goal. (a) EmbCLIP deviates from the optimal route; (b) Our method directly navigates to the target point by choosing an optimal route.}
  \label{fig:case_eg_habitat}
\end{figure}


\subsection{Results}
\label{exp_rst}

Table~\ref{tab:rst_objnav_robothor} presents our experimental results for each task. As discussed in Section~\ref{exp_setting}, we compare our method with two types of approaches, thus our findings align accordingly:  

1) Compared to EmbCLIP, which shares the same model architecture and training settings, our method significantly outperforms the baseline across all tasks and evaluation metrics. Notably, it nearly doubles EmbCLIP's performance in RoboTHOR and Habitat ObjNav, as measured by SPL and SR, demonstrating its effectiveness and efficiency.  

2) Compared to methods that rely on stronger assumptions and inductive biases, our approach consistently outperforms them, albeit with a smaller margin in some cases. This further highlights the robustness of our method and its ability to achieve superior results without dataset-specific designs.

Additionally, we conducted \textbf{case studies} to examine the impact of our method. In RoboTHOR ObjNav (Figure~\ref{fig:case_eg}), we tested on the task \textit{Find an AlarmClock} and found that our approach led the agent to stop closer to the objective, improving performance. In Habitat PointNav (Figure~\ref{fig:case_eg_habitat}), we evaluated the methods in the \textit{Nuevo} scenario, where the green dot marks the start and the red star indicates the target. Our agent navigated directly along an optimal route, whereas it failed to find the shortest path without the Causal Understanding Module.  

In summary, both quantitative results and the qualitative sample check demonstrated the effectiveness and generalizability of our proposed method.

\subsection{Ablation Studies}
\label{exp_ablation}
We conduct ablation studies to examine the effectiveness of each component in this Section.

\begin{figure}
    \begin{center}
        \includegraphics[width=0.95\linewidth]{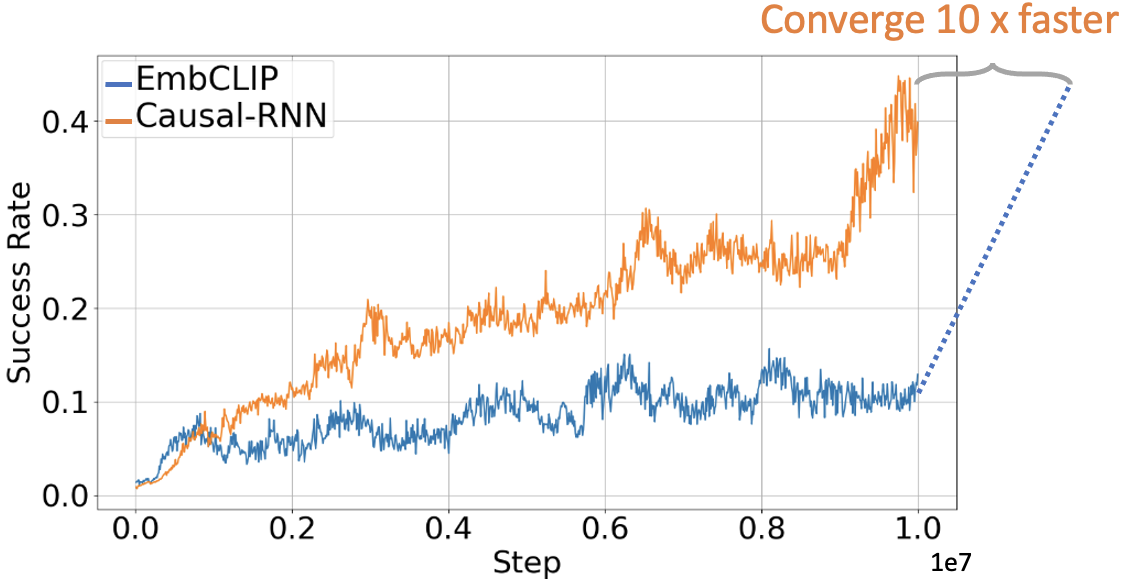}
    \end{center}
    \caption{Average Success Rate for EmbCLIP and Causal-RNN on RoboTHOR ObjNav. Our proposed Causal Understanding Module can: 1) significantly benefit the performance; 2) significantly reduce the training time by 10 times.}
    \label{fig:causalrnn_sr_curve}
\end{figure}

\textbf{Removing Causal} We examine the impact of the Causal Understanding Module by removing it from the architecture while keeping all other settings unchanged. We refer to this model as \textbf{Transformer} in Table~\ref{tab:ablation_rst} for simplicity.

\textbf{Removing Transformer} We examine the impact of the Transformer Encoder by replacing it with an RNN while keeping all other settings unchanged, i.e., this is equivalent to implementing the Causal Understanding Module on EmbCLIP \cite{khandelwal2022simple}. We refer to this model as \textbf{Causal-RNN} for simplicity.

\textbf{Results} The results are provided in Table~\ref{tab:ablation_rst}. We observe that while a Multi-modal Transformer Encoder contributes to the agent's performance, the improvement is limited. In contrast, the Causal Understanding Module significantly improves the model's performance, and \textbf{most of the observed improvements can be attributed to this module}.

\textbf{Curve of Success Rate} To further investigate the impact of the Causal Understanding Module, we plot the Success Rate curve of Causal-RNN and EmbCLIP (which differ only in the Causal Understanding Module) in Figure~\ref{fig:causalrnn_sr_curve}. The values represent the Average Success Rate over 10 random runs. We observe \textbf{two key benefits} of the module: 1) It significantly enhances the baseline model's performance without computational overhead, increasing the success rate by over 50\% with just a simple linear layer; 2) It significantly reduces training time. Specifically, Causal-RNN achieves a success rate of 0.48 within 15M steps, whereas EmbCLIP, as reported in \cite{khandelwal2022simple}, reaches 0.47 only after 200M steps. Thus, our approach reduces training time by \textbf{over 10 times} with a minimal architectural modification.

\subsection{Causal Understanding Module in Supervised Learning}
\label{exp_other_setting}

\begin{table}
  \centering
  \caption{Our method is broadly applicable and benefitial to various baseline methods in Supervised Learning.}
  \begin{tabular}{l | l l l}
    \toprule
    & NE $\downarrow$ & OSR $\uparrow$ & SR $\uparrow$\\
    \midrule
    Seq2Seq & 7.8 & 28.4 & 21.8 \\
    \:\:\: + Ours & 5.8 \textcolor{teal}{$\downarrow$(-2.0)} & 38.1 \textcolor{teal}{$\uparrow$(+9.7)} & 41.3 \textcolor{teal}{$\uparrow$(+19.5)} \\
    \midrule
    Speaker Follower & 6.6 & 45.0 & 35.0 \\
    \:\:\: + Ours  & 4.9 \textcolor{teal}{$\downarrow$(-1.7)} & 53.1 \textcolor{teal}{$\uparrow$(+8.1)} & 52.4 \textcolor{teal}{$\uparrow$(+17.4)} \\
    \midrule
    EnvDrop & 5.2 & - & 52.2 \\
    \:\:\: + Ours & 4.5 \textcolor{teal}{$\downarrow$(-0.7)} & - & 68.7 \textcolor{teal}{$\uparrow$(+16.5)} \\
    \bottomrule
  \end{tabular}
  \label{tab:rst_r2r}
\end{table}

While the earlier sections focus on experiments in the Reinforcement Learning setting, we also evaluate our method in supervised learning. Specifically, we integrate the Causal Understanding Module into existing methods, including Seq2Seq \cite{anderson2018vision}, Speaker Follower \cite{fried2018speaker}, and EnvDrop \cite{tan2019learning}, while keeping all other settings unchanged. Experiments are conducted on the R2R dataset \cite{anderson2018vision}, and performance is assessed using standard metrics such as Navigation Error (NE), Oracle Success Rate (OSR), and SR, as introduced earlier. Table~\ref{tab:rst_r2r} presents the results on the validation unseen split, demonstrating the effectiveness, consistency, and generalizability of the Causal Understanding Module.

\section{Related Work}
\subsection{Vision-based Navigation}
Navigation, as a fundamental task in Embodied AI \cite{anderson2018vision,ku2020room,qi2020reverie}, has inspired numerous methods aimed at improving action planning. Existing approaches incorporate techniques such as historical state tracking \cite{chen2021history,feng2023gpf}, navigation map generation \cite{chen2021topological,chen2022think,wang2023gridmm,zhao2022target}, and external knowledge prompts \cite{li2023kerm,zhao2023zero,guan2024loc,xu2024aligning}. While these methods achieve strong performance, they are often tailored to specific tasks \cite{inoue2022prompter, blukis2022persistent} and thus lack generalizability \cite{duan2022survey}. \cite{khandelwal2022simple} addressed this issue by proposing a baseline without inductive biases.

With the rise of pre-trained Large Foundation Models, recent studies have explored integrating pre-trained knowledge into VLN. Some works \cite{zhou2024navgpt,long2024discuss,chen2024mapgpt,cai2024bridging} demonstrate the zero-shot potential of off-the-shelf LLMs for navigation. However, despite leveraging advanced models such as GPT-4 \cite{achiam2023gpt}, these approaches still lag behind supervised methods. To bridge this gap, other studies \cite{pan2023langnav,lin2024navcot,zheng2024towards} fine-tune LLaMA-7B \cite{touvron2023llama} on VLN-specific datasets, assessing language as a perceptual tool for navigation.

\subsection{Causality in Embodied AI}
Causality has gained increasing attention in machine learning tasks \cite{yue2020interventional,kocaoglu2017causalgan}. In Embodied AI, it has been applied to Reinforcement Learning \cite{deng2023causal} and Imitation Learning \cite{de2019causal}. \cite{gupta2024essential} highlights the essential role of causality in Embodied AI, while \cite{zhang2020learning} models latent transition dynamics using causal core sets. \cite{li2020causal} addresses confounding effects in state transition modeling. Other works \cite{annabi2022intrinsically,volodin2020resolving,zhu2022offline,wang2022causal,tomar2021model} construct SEMs among state variables, ensuring transition dynamics depend only on causally relevant features. Recently, \cite{richens2024robust} proved that agents capable of generalizing under domain shifts must learn a causal model.


\section{Conclusion}
We address key challenges in Navigation, particularly the limitations of prevalent vanilla sequential data modeling methods. By highlighting the intrinsic differences between Navigation tasks and traditional sequential data tasks, we introduce a novel causal framework that justifies the need for a causal environment understanding module. We propose Causality-Aware Navigation (CAN), an end-to-end transformer-based method that shows significant performance improvements across various tasks and simulators. Ablation studies demonstrate that the majority of performance gains can be attributed to the Causal Understanding Module, which is proven effective and can be incorporated into other methods in both Reinforcement Learning and Supervised Learning without computational overhead. This work advances agent capabilities through a generalizable causal approach, opening up potential for broader applications.



\bibliographystyle{IEEETran}
\bibliography{reference}

\end{document}